\documentclass{article} 
\usepackage{iclr2023_conference,times}

\usepackage{amsmath,amsfonts,bm}

\def\eqref#1{equation~\ref{#1}}

\def\1{\bm{1}}

\DeclareMathAlphabet{\mathsfit}{\encodingdefault}{\sfdefault}{m}{sl}
\SetMathAlphabet{\mathsfit}{bold}{\encodingdefault}{\sfdefault}{bx}{n}

\DeclareMathOperator*{\argmax}{arg\,max}

\usepackage{hyperref}
\usepackage{url}

\usepackage{graphics} 
\usepackage{caption}
\usepackage{subcaption}
\usepackage{epsfig} 
\usepackage{mathptmx} 

\usepackage{multirow}
\usepackage{makecell}
\usepackage{amsmath}
\usepackage{textcomp}
\usepackage{xcolor}
\usepackage{algorithm}
\usepackage{algorithmicx}
\usepackage{comment}
\usepackage{algpseudocode}

\usepackage{footnote}
\usepackage[bottom]{footmisc}
\usepackage{threeparttable}
\usepackage{multirow}
\usepackage{tabularx}
\usepackage{ragged2e}
\usepackage{booktabs}
\usepackage{makecell}
\usepackage{hyperref}
\usepackage{xspace}

\usepackage{soul}

\newcolumntype{C}[1]{>{\centering\arraybackslash}p{#1}}
\newcolumntype{P}[1]{>{\centering\arraybackslash}p{#1}}
\newcolumntype{M}[1]{>{\centering\arraybackslash}m{#1}}
\newcolumntype{L}[1]{>{\raggedright\arraybackslash}p{#1}}
\newcolumntype{R}[1]{>{\raggedleft\arraybackslash}p{#1}}
\newcolumntype{J}[1]{>{\justifying\arraybackslash}p{#1}}

\usepackage[english]{babel}
\usepackage{amsthm}
\theoremstyle{definition}

\newtheorem*{assumption*}{\assumptionnumber}
\providecommand{\assumptionnumber}{}


\usepackage{footnote}
\usepackage[bottom]{footmisc}
\usepackage{threeparttable}
\usepackage{tabularx}
\usepackage{ragged2e}
\usepackage{hyperref}
\newcolumntype{C}[1]{>{\centering\arraybackslash}p{#1}}
\newcolumntype{P}[1]{>{\centering\arraybackslash}p{#1}}
\newcolumntype{M}[1]{>{\centering\arraybackslash}m{#1}}
\newcolumntype{L}[1]{>{\raggedright\arraybackslash}p{#1}}
\newcolumntype{R}[1]{>{\raggedleft\arraybackslash}p{#1}}
\newcolumntype{J}[1]{>{\justifying\arraybackslash}p{#1}}

\usepackage{xspace}
\newcommand{\mypara}[1]{\noindent\textbf{#1}}

\newcommand{\myResFed}{\emph{ResFed}\xspace}

\definecolor{amethyst}{rgb}{0.6, 0.4, 0.8}
\definecolor{aqua}{rgb}{0.0, 1.0, 1.0}
\definecolor{arylideyellow}{rgb}{0.91, 0.84, 0.42}
\definecolor{phlox}{rgb}{0.87, 0.0, 1.0}
\definecolor{caribbeangreen}{rgb}{0.0, 0.8, 0.6}

\title{ResFed: Communication Efficient Federated Learning by Transmitting Deep Compressed Residuals}

\iclrfinalcopy

\author{
  \textbf{Rui Song\:$^1$$^,$$^2$}
  \textbf{Liguo Zhou\:$^2$}
  \textbf{Lingjuan Lyu\:$^3$}
  \textbf{Andreas Festag\:$^1$$^,$$^4$}
  \textbf{Alois Knoll\:$^2$}
  \\
  $^1$ Fraunhofer IVI, Germany \\ 
  $^2$ Technical University of Munich, Germany \\
  $^3$ Sony AI, Japan \\
  $^4$ Technische Hochschule Ingolstadt, Germany \\
  \texttt{\{rui.song, andreas.festag\}@ivi.fraunhofer.de}
  \\
  \texttt{\{rui.song, liguo.zhou\}@tum.de,  knoll@in.tum.de}
  \\
  \texttt{lingjuan.lv@sony.com}
  \\
  \texttt{andreas.festag@thi.de} 
}

\begin{document}

\maketitle

\begin{abstract}
Federated learning enables cooperative training among massively distributed clients by sharing their learned local model parameters.
However, with increasing model size, deploying federated learning requires a large communication bandwidth, which limits its deployment in wireless networks. 
To address this bottleneck, we introduce a residual-based federated learning framework (ResFed), where residuals rather than model parameters are transmitted in communication networks for training. 
In particular, we integrate two pairs of shared predictors for the model prediction in both server-to-client and client-to-server communication.
By employing a common prediction rule, both locally and globally updated models are always fully recoverable in clients and the server. 
We highlight that the residuals only indicate the quasi-update of a model in a single inter-round, 
and hence contain more dense information and have a lower entropy than the model, comparing to model weights and gradients.
Based on this property, we further conduct lossy compression of the residuals by sparsification and quantization and encode them for efficient communication.
The experimental evaluation shows that our ResFed needs remarkably less communication costs and achieves better accuracy by leveraging less sensitive residuals, compared to standard federated learning.
For instance, to train a 4.08~MB CNN model on CIFAR-10 with 10~clients under non-independent and identically distributed (Non-IID) setting, our approach achieves a compression ratio over $700\times$ in each communication round with minimum impact on the accuracy. 
To reach an accuracy of 70\%, it saves around 99\% of the total communication volume from 587.61~Mb to 6.79~Mb in up-streaming and to 4.61~Mb in down-streaming on average for all clients.
\end{abstract}

\section{Introduction}
\label{sec:intro}


\begin{figure*}[!ht]
\centering
\begin{subfigure}[b]{1\textwidth}
   \includegraphics[trim=0 0 0 0,clip,width=1\linewidth]{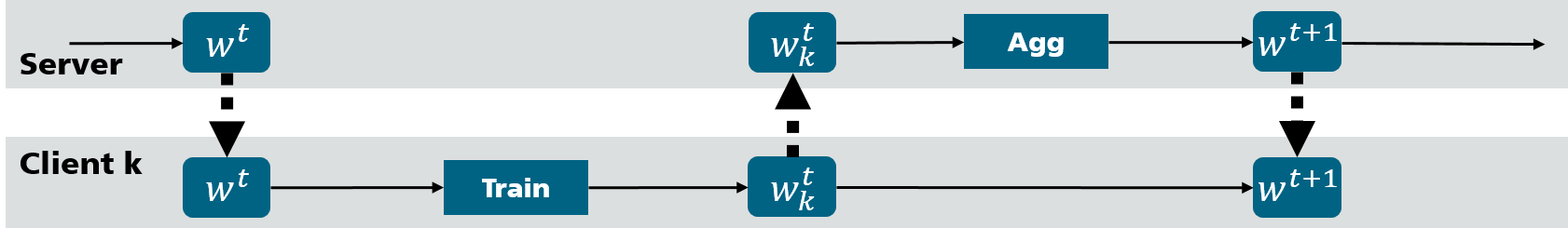}
   \caption{Standard federated learning system}
   \label{fig:ResFed_cap1} 
\end{subfigure}
\begin{subfigure}[b]{1\textwidth}
  \includegraphics[trim=0 0 0 0,clip,width=1\linewidth]{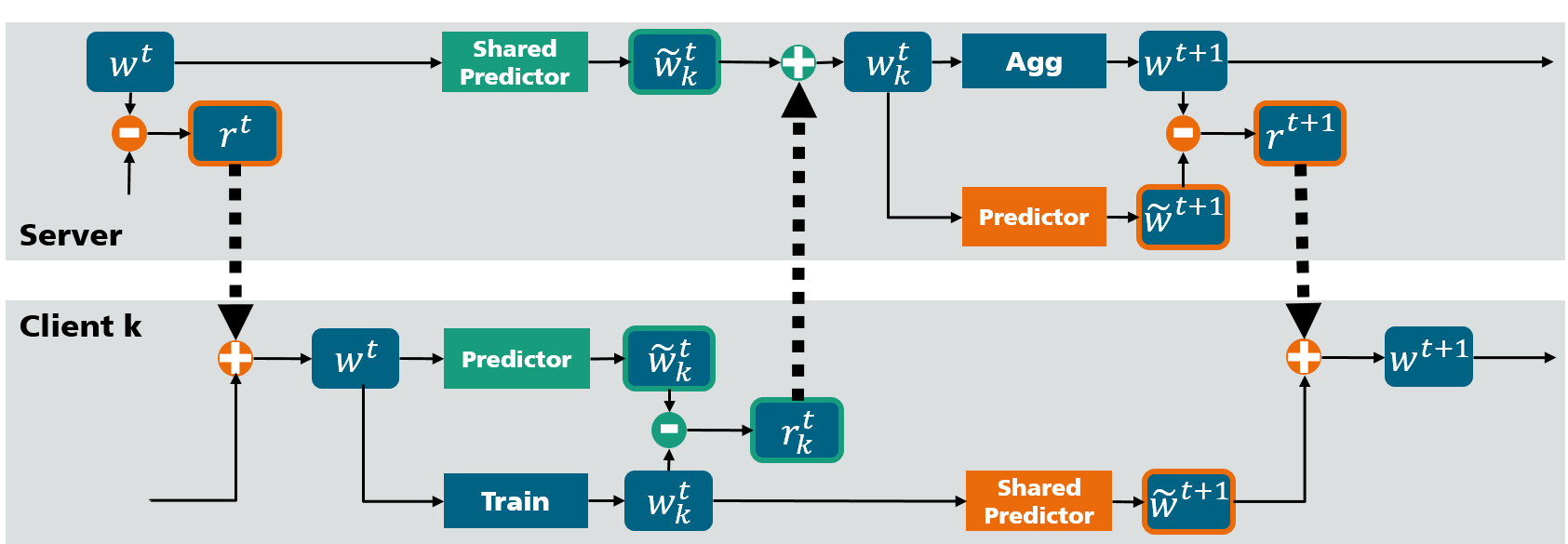}
   \caption{\centering Residual-based federated learning system}
   \label{fig:ResFed_cap2}
\end{subfigure}
\caption{Paradigm shift from standard federated learning to residual-based federated learning system, with additional two pairs of predictors and corresponding operators (in green and orange). The model is updated in the client $k$ by local training (Train) and in the server by aggregation (Agg).}
\label{fig:ResFed}
\end{figure*}

Federated learning has become an emerged machine learning paradigm, which enables distributed training on broad data sources without disclosing their original data~\cite{pmlr-v54-mcmahan17a}. 
Instead of transmitting raw data, only parameters (mostly model weights or gradients) in federated learning are iteratively shared between clients and a server via heterogeneous networks. 
Federated learning has been successfully applied to various applications~\cite{lyu2020privacy}, such as mobile keyboard prediction, speech recognition, image object detection, etc, 
However, with the increasing size of machine learning models, the existing mobile communication infrastructure cannot always meet the requirement in terms of bandwidth and latency in federated learning, which constraints the wide deployment of federated learning. 
For instance, to train a transformer model with billions of parameters usually 32-bit float parameters), the size of a message in a single federated learning round can be several 10 or 100~Gigabytes, e.g. a CTRL model~(\cite{keskar2019ctrl}) with 1.6 billions parameters or a T5 model~(\cite{2020t5}) with up to 11 billions parameters.
That can cause an enormous and extremely costly data traffic, even in 5G NR networks, where the throughput can be from 5~Gbps to 18~Gbps.
Another application scenario is to improve machine learning models for road traffic object recognition and detection in V2X (Vehicle-to-Everything) communication networks, where the bandwidth for V2X is also occupied for other traffic services at the same time, e.g. collective perception service, and obviously the safety-related services should have higher priority. 
Therefore, communication efficiency is a pivotal component for deploying federated learning, especially in wireless networks.
In an attempt to tackle the communication bottleneck, the parameter compression is considered as one of the most effective approaches, which allows for updating the models by transmitting much smaller size of messages in networks, and thereby reduces the required time per communication round in federated learning.
The approaches proposed by~\cite{xu2020ternary, reisizadeh2020fedpaq, honig2022dadaquant} can effectively reduce the communication volume in each round by various quantization techniques, however they only consider the 
communication efficiency for uploading (client-to-server) but not for downloading (server-to-client). 
\cite{lin2018deep} compress the gradients instead of model parameters for distributed learning, which can not well fit federated learning, where clients can train multiple epochs in each round. 
\cite{wu2022communication} use knowledge distillation (\cite{hinton2015distilling}) to learn and transmit a smaller model, where the original model structure is affected. 
Furthermore, all of those works attempt to compress the model parameters or gradients based on the model in a specific round, without consideration of inter-round model update similarity, which contains additional redundancy sequentially.

Inspired by residuals in video compression protocols from~\cite{li2021deep}, we introduce a residual-based federated learning framework, termed as \myResFed. 
It allows the server and clients to share and update models by sharing model parameter residuals rather than model parameters or gradients.
Particularly, by observing training trajectory in each local client and the aggregation trajectory in the server, we believe model updates in both clients and the server can be predictable. 
Those predictive models -- in analogy with predictive frames in video transmission -- can foresee model updates in the federated learning. 
After each communication round, we use the deviations between the predictive and the actual updated model parameters, which we call the model residuals, 
for the communication in networks.
Note that the actual updated models can be always recoverable by acquiring the residuals, as the predictors in the senders are shared to the receivers in \myResFed. More details are provided in Sec.~\ref{sec:method} and~\ref{sec:resfed}.

Unlike transmitting model weights, \myResFed can wring out the potential redundancy by removing the predictive information from history updates and only keep the residuals for communication. Compared to transmitting model gradients after each training epoch, \myResFed allows the models to be trained locally multiple times. Compared to transmitting residual accumulation for multiple epochs, \myResFed further minimizes the information by predicting the model updates from history. As shown in Fig.\ref{fig:param_res}, the values of residuals are overall smaller than weights and gradients during the entire training process. To further shrink the size of messages for communication, we then compress only residuals using sparisification and quantization, and encode the messages for information sharing in client-to-server and server-to-client.

Our main contributions are summarized as follows:

\begin{itemize}
    \item We introduce and formulate the model residuals for the communication efficiency in federated learning and indicate the residuals contain more dense information than model weights and gradients.
    \item We propose a novel federated learning framework  (\myResFed) based on deep residual compression, which consists of the following steps: predictor sharing, model prediction, residual generation, residual compression, residual communicating, model recovering and model trajectory synchronization. 
    \item We provide the experimental evaluation of our framework with various communication cost budgets in both up- and down-streaming, which gives an insight in deploying it in resource-constrained communication environments. The open source implementation of \myResFed will be publicly available.
\end{itemize}

\section{System Setup}
\label{sec:method}

We first introduce the related concepts and techniques that will be used in our framework. Given $N$ clients and a server in a federated learning system, we only focus on the information sharing between one single client $k$ and the server from communication round $t$ to $t+1$, as shown in Fig~\ref{fig:ResFed}. The information sharing for other clients is the same.

\subsection{Model Update}
\label{sec:update}

\mypara{Client.}
Given a client $k$ with a local dataset $\mathcal{D}_k$, the initial local model in a new round $t$ is $w_k^{t-1}$. Before the local training starts, the client initially receives the global model $w^t$ from the server and updates the local model to $\hat w_{k}^t$.
After that, local model $\hat w_{k}^t$ is trained on $\mathcal{D}_k$ and transited to $w_{k}^t$. We mark the first update as $w_{k}^{t-1} \rightarrow \hat w_{k}^t$ and the second one as $\hat w_{k}^t \rightarrow w_{k}^t$.
Note that the first updated model is equal or similar to the global updated model $w^t$, i.e. $\hat w_{k}^t \simeq w^t$. If lossy compression is used for communication and the loss due to compression can not be repaired, then $\hat w_{k}^t \sim w^t$.

\mypara{Server.}
Similarly, the global model $w_{t-1}$ in the server is also updated twice after one round of communication $t$. The first update happens when it receives models from the clients, i.e. $w^{t-1} \rightarrow \{\hat w_i^t|i=1,2,...,N\}$. Then the aggregation leads to the second update, $\{\hat w_i^t|i=1,2,...,N\}\rightarrow w^{t}$.

\subsection{Model Trajectory}
\label{sec:trajectory}
\mypara{Client.}
Given a client $k$ at time point $t$, we cache the updated models with a sliding time window $[t-T, t]$ in two different queues, that distinguish by two model updates. 
We refer the time sequence of local model updates $\mathcal{L}_{k}^t=\{w_{k}^{t-T}, ...,w_{k}^t\}$ from $w_{k}^t \rightarrow \hat w_{k}^t$ as a local model trajectory, and $\mathcal{\hat G}_{k}^t=\{\hat w_{k}^{t-T}, ...,\hat w_{k}^t\}$ from $\hat w_{k}^t \rightarrow w_{k}^{t+1}$ as a global model trajectory.

\mypara{Server.}
Correspondingly, we cache the local and global model updates in the local and global model trajectories for all client at the server, i.e. $\{\mathcal{\hat L}_i|i=1,2,...,N\}$ and $\{\mathcal{G}_i|i=1,2,...,N\}$. Note that if the server can always send the lossless global model update to all clients, the global trajectories at time $t$ are the same for all clients.

\subsection{Model Prediction}
\label{sec:prediction}
\mypara{Client.}
Given a client $k$ at time point $t$, we predict $\hat w_{k}^t \rightarrow w_{k}^t$ from the local and global training trajectories, $\mathcal{L}_{k}^{t-1}$ and $\mathcal{\hat G}_{k}^{t-1}$ as follows:
\begin{equation}
\label{eq:prediction_client}
    \tilde w_{k}^{t} = f_{predict,k}(\mathcal{L}_{k}^{t-1}, \mathcal{\hat G}_{k}^{t-1}, \hat w_{k}^{t}) = \argmax_{w_{k}^t} p(w_{k}^t|\underbrace{w_{k}^{t-T}, ...,w_{k}^{t-1}}_{\text{$\mathcal{L}_{k}^{t-1}$}},\underbrace{\hat w_{k}^{t-T}, ...,\hat w_{k}^{t-1}}_{\text{$\mathcal{\hat G}_{k}^{t-1}$}}, \hat w_{k}^{t})
\end{equation}
where $f_{predict,k}$ is the used predictor for model prediction in the client $k$.

\mypara{Server.}
For the server, we predict model updates $\hat w_i^t\rightarrow w^t$ for each client $i$ from local and global trajectories $\mathcal{\hat L}_{i}^{t-1}$ and $\mathcal{G}_{i}^{t-1}$ as follows:
\begin{equation}
\label{eq:prediction_server}
    \tilde w_{i}^{t} = h_{predict,i}(\mathcal{\hat L}_{i}^{t-1}, \mathcal{G}_{i}^{t-1}, \hat w_{i}^{t}) = \argmax_{w_{i}^t} p(w_{i}^t|\underbrace{w_{i}^{t-T}, ...,w_{i}^{t-1}}_{\text{$\mathcal{\hat L}_{k}^{t-1}$}},\underbrace{\hat w_{i}^{t-T}, ...,\hat w_{i}^{t-1}}_{\text{$\mathcal{G}_{k}^{t-1}$}}, \hat w_{i}^{t}), \forall i\in {1,...,N}
\end{equation}
where $h_{predict,i}$ is the used predictor for model prediction in the server for each client $i$.

\subsection{Model Residual}
\label{sec:residual}

Given a model update $\hat w_{i}^t \rightarrow w_{i}^t$ for the client $i$ at time $t$ in the server or in the corresponding client, if we can compute the model prediction $\tilde w_{i}^{t}$ based on Eq.~\ref{eq:prediction_client}, we define the model residual as follows:
\begin{equation}
\label{eq:residual}
    r_{i}^t =  w_{i}^t - \tilde w_{i}^t
\end{equation}
Note that $r_{i, ul}^t$ and $r_{i, dl}^t$ are the residuals of clients for uploading and the residuals of the server for downloading, respectively.More understanding of residuals is provided in Sec.~\ref{sec:understanding}.

\subsection{Deep Compression}
\label{sec:compression}

\label{eq:compression}
To reduce the model size for more efficient communication, we shrink the model size before sending it out. We define a compressed model in the client k: 
\begin{equation}
   \bar w_{k}^t = f_{compress}(w_{k}^t)
\end{equation}
and in the server:
\begin{equation}
    \bar w_{i}^t = h_{compress}(w_{i}^t), \forall i\in \{1,...,N\}
\end{equation}
where $f_{compress}$ and $h_{compress}$ is the used compressor for model compression in clients and server respectively. In our system, we consider to compress and communicate model residuals instead of model itself in the client k: 

\begin{equation}
    \label{eq:r_dl}
    \bar r_{k,ul}^t = f_{compress}(r_{k,ul}^t) = f_{compress}( w_{k}^t -  f_{predict, i}(w_{k}^{t-T}, ...,w_{i}^{t-1},\hat w_{k}^{t-T}, ...,\hat w_{k}^{t-1}, \hat w_{k}^{t}))
\end{equation}
and in the server:
\begin{equation}
    \label{eq:r_ul}
    \bar r_{i,dl}^t = h_{compress}(r_{i,dl}^t) = h_{compress}( w_{i}^t -  f_{predict, i}(w_{i}^{t-T}, ...,w_{i}^{t-1},\hat w_{i}^{t-T}, ...,\hat w_{i}^{t-1}, \hat w_{i}^{t})), \forall i\in \{1,...,N\}
\end{equation}

\section{ResFed: Residual-based Federated Learning Framework}
\label{sec:resfed}

\begin{figure*}[t]
\includegraphics[trim=0 0 0 0,clip,width=1\linewidth]{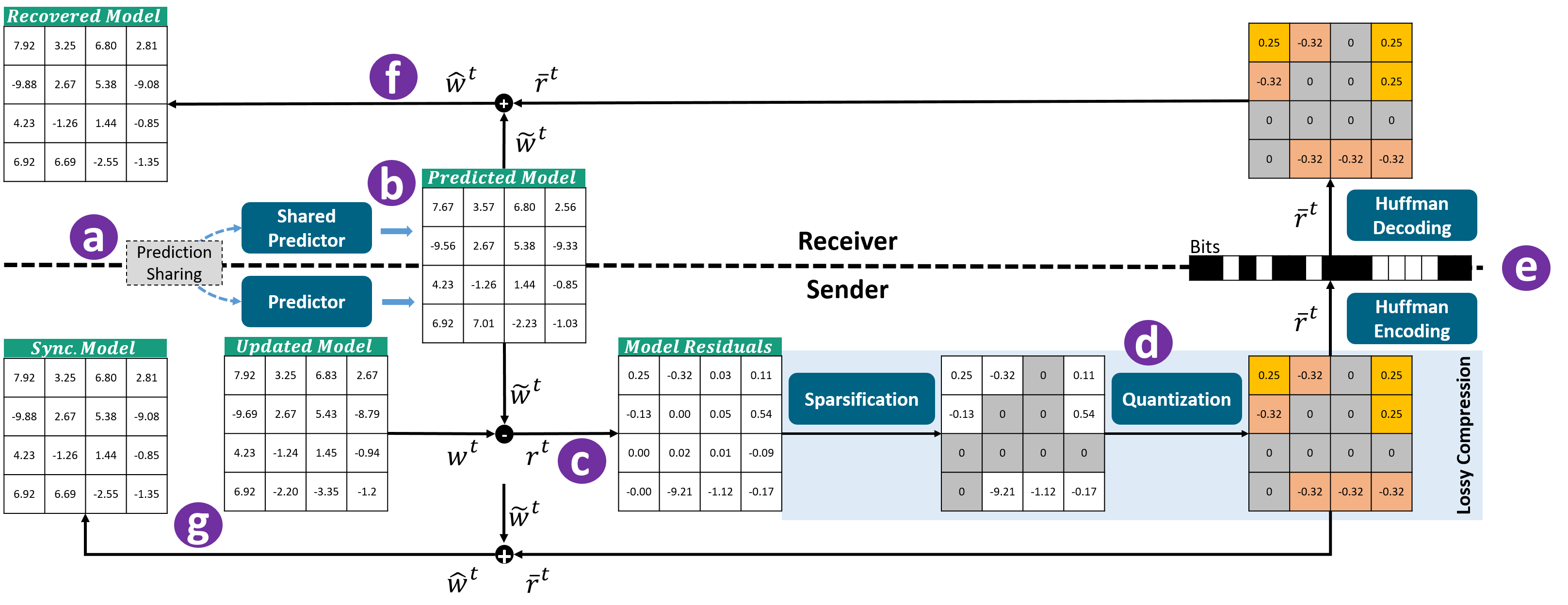}
\caption{
\myResFed system model with a lossy deep compression pipeline for efficiently transmitting encoded model residuals. The following steps should be implemented for one communication: \emph{(a)}~Share the predictor for both sender and receiver before the federated learning starts; \emph{(b)} 
Execute the same model prediction before communicating any model information; \emph{(c)} Generate the model residuals; \emph{(d)}~Compress residuals using deep compression; \emph{(e)}~Communicate residual bits with encoding and decoding; \emph{(f)}~Recover the model from received residuals; \emph{(g)}~Synchronize the model trajectory by simulation, recovering the model locally with consideration of lossy compression.}
\vspace{0cm}
\label{fig:lossyResFed}
\end{figure*}

\begin{algorithm}[t]
 \caption{\raggedright: \myResFed: Residual-based federated learning framework}
 \label{alg:FedRes}
 \begin{algorithmic}[1]
    \State \textbf{Server runs:} 
    \State initialize the global model $w$
    \State initialize the empty local model trajectories $\mathcal{\hat L}_{1},...,\mathcal{\hat L}_{N}$ 
    \State initialize the global model trajectories $\mathcal{G}_{1},...,\mathcal{G}_{N}$ 
    \State initialize the predictor $h_{predict}$
    \For{$i \in \{1,2,...,N\}$}
        \State initialize an empty local model trajectories $\mathcal{L}_{i}$ \Comment{@client $i$}
        \State initialize an empty global model trajectories $\mathcal{\hat G}_{i}$ \Comment{@client $i$}
        \State $h'_{predict,i} = h_{predict}$ \Comment{sharing predictors to client $i$}
        \State $f'_{predict,i} = f_{predict, i}$ \Comment{get the shared predictors from client $i$}
    \EndFor
    \For{$t \in \{1,2,...,M\}$}
        \For{$i \in \{1,2,...,N\}$ \textbf{in parallel}}
            \If{$t < T$}
                \State $\mathcal{\hat G}_i \leftarrow cache(w)$ \Comment{cache the global model in $\mathcal{\hat G}_i $ @client $i$}
                \State Server communicates $w$ to the client $i$
                \State $\hat w_i \leftarrow \textbf{LocalTrain}(w)$ \Comment{@client $i$}
                \State Client $i$ communicates $\hat w_i$ to the server
                \State $\mathcal{L}_i \leftarrow cache(\hat w_i)$ \Comment{cache the local model in $\mathcal{L}_i$ @client $i$}
            \Else
            \State Server communicates $\bar r_{i,dl}$ to the client $i$
            \State $\bar r_{i,ul} \leftarrow$
            \textbf{ResFedClientUpdate} $(i, \bar r_{i,dl})$ \Comment{@client $i$}
            \State Client $i$ communicates $\bar r_{i,ul}$ to the server
            \State $\tilde w_i \leftarrow f'_{predict,i}(\mathcal{G}_i, \mathcal{\hat L}_i, \hat w)$ \Comment{predict updated model}
            \State $\hat w_i \leftarrow \tilde w_i + \bar r_{i,ul}$ \Comment{recover models}
            \EndIf
         \State $\mathcal{\hat L}_i \leftarrow cache(\hat w_i)$  \Comment{update local trajectory} 
        \EndFor
        \State $w \leftarrow \textbf{Aggregate}(\hat w_1,...,\hat w_N)$
        \For{$i \in \{1,2,...,N\}$ }
            \State $\tilde w_i \leftarrow h_{predict}(\mathcal{G}_i, \mathcal{\hat L}_i, w)$ \Comment{predict updated model}
            \State $r_{i,dl} \leftarrow  w - \tilde w_i$ \Comment{compute model residuals}
            \State $\bar r_{i,dl} \leftarrow h_{compress}(r_{i,dl})$ \Comment{compress model residuals}
            \State $\mathcal{G}_i \leftarrow cache(\tilde w_i + \bar r_{i,dl})$ \Comment{synchronize global trajectory}
        \EndFor
    \EndFor
    \State \textbf{return} $w$ 
    \vspace{0.2cm}
    \State \textbf{ResFedClientUpdate} $(k, \bar r)$
    \State $\tilde w \leftarrow h'_{predict, i}(\mathcal{\hat G}_k, \mathcal{L}_k, w_k)$ \Comment{predict updated model}
    \State $\hat w \leftarrow \tilde w + \bar r$ \Comment{recover models}
    \State $\mathcal{\hat G}_k \leftarrow cache(\hat w)$ \Comment{update global trajectory}
    \State $w_k \leftarrow \textbf{LocalTrain}(\hat w)$
    \State $\tilde w_k \leftarrow f_{predict,i}(\mathcal{\hat G}_k, \mathcal{L}_k, \hat w)$ \Comment{predict updated model}
    \State $r_k \leftarrow  w_k - \tilde w_k$ \Comment{compute model residuals}
    \State $\bar r_k \leftarrow f_{compress}(r_k)$ \Comment{compress model residuals}
    \State $\mathcal{L}_k \leftarrow cache(\tilde w_k + \bar r_k)$ \Comment{synchronize local trajectory}
    \State \textbf{return} $\bar r_k$
\end{algorithmic}
\end{algorithm}

The overview of the \myResFed is shown in Fig.~\ref{fig:ResFed} and the detailed steps with lossy compression in one communication round 
is given in Fig.~\ref{fig:lossyResFed}. In particular, we introduce \emph{(a)}~predictor sharing, \emph{(b)}~model prediction, and \emph{(c)}~residual generation in Sec~\ref{sec:sharing}. In Sec.~\ref{sec:compression}, \emph{(d)}~residual compression is formulated in Eq.~\ref{eq:r_dl} and Eq.~\ref{eq:r_ul}. Then, after \emph{(e)}~communicating residual bits, we provide details on \emph{(f)}~model recovering in Sec.~\ref{sec:recovery}. Finally in Sec. ~\ref{sec:sync}, we describe \emph{(g)}~trajectory synchronization.

\subsection{Predictor Sharing}
\label{sec:sharing}

Then, we consider to deploy a pair of predictors in both clients and the server, which can 
execute the model predictions based on the local and global model trajectories in the time series. In our framework, the server 
caches the local model trajectories in all clients once the local model update is received. Given a client $k$, if the received models in the server is exactly the same as the local model, and the models in $\mathcal{\hat G}_k$ also the same, we say that both trajectories in client $k$ are fully observable in the server.

Give a predictor $f_{predict, k}$ in the client $k$, we share it to make the predictor in the server $f'_{predict, k}=f_{predict, k}$ in \myResFed. If the trajectories in client are fully observable in the server, we can get the same model predictions in both server and clients from Eq.~\ref{eq:prediction_client}. Then, by communicating the model residuals, the new model update at time $t+1$ can be recovered by the model residual and the model at last time point. Also, we share the set of predictors in the server $\{h_{predict, i}|i=1,...,N\}$ to the corresponding client, i.e. $h'_{predict, i}=h_{predict, i}, \forall i\in \{1,...,N\}$. Then we can get the same model predictions based on Eq.~\ref{eq:prediction_server}.

For the predictor, various design choices exist. In this work, we employ the predictor with respect to the model transition dynamics in a sliding history time window. The predictor is formulated as follows:

\begin{equation}
    \tilde w_{k}^{t} = f_{predict}(\mathcal{L}_{k}^{t-1}, \mathcal{\hat G}_{k}^{t-1}, \hat w_{k}^t) =
    \begin{cases} 
    \hat w_{k}^t, & \text{$T=0$}.\\
    \hat w_{k}^t + \sum_{\tau=1}^T (-1)^{T-\tau}(T-\tau+1)(w_{k}^{t-\tau}-\hat w_{k}^{t-\tau}), & \text{$T>0$}.
    \end{cases}
    \label{prediction}
\end{equation}

To reduce the used memory for caching trajectories in the client, we apply a short time window $[t-T, t]$ in the prediction process. We term \emph{(i)} \emph{stationary predictor} when $T=0$; and \emph{(ii)} \emph{linear predictor} when $T=1$. Note that we consider the model updates in~\cite{barnes2020rtop,mitchell2022optimizing,isik2022information} as special residuals, which calculated by stationary predictor.

Specifically, the stationary predictor uses the current model for the prediction of the next model, $\tilde w_{k}^{t} = \hat w_{k}^t$, where the model residual is always $r_{k}^{t} = w_{k}^{t} - \hat w_{k}^{t}$. 
Note that when the number of local training epochs is fixed to 1, the stationary residuals is proportional to gradients, i.e. $r = \eta g$, where $\eta$ is the learning rate and $g$ represents the gradients.
In the linear predictor, $\tilde w_{k}^{t} = \hat w_{k}^{t} + w_{k}^{t-1} - \hat w_{k}^{t-1}$, the model transition in the last local training step is always considered, where $r_{k}^{t} = w_{k}^{t} - \hat w_{k}^{t} - w_{k}^{t-1} + \hat w_{k}^{t-1}$. 
The predictor for the client $k$ in the server $h_{predict,k}$ is similar to $f_{predict,k}$.

\subsection{Model Recovery}
\label{sec:recovery}

We cache the trajectories in both server and clients. 
Each client has two model trajectories for local and global model updates in the history. %
In the server, it caches the global trajectory and the local model trajectories of all connecting clients in the history. 
In this case, the two trajectories in each client are fully observable in the server. Through sharing predictors, given a client $k$ at round $t$, the server is able to 
get the same model prediction $\tilde w_{k}^t$ as the client $k$. 

If uncompressed model residuals ($\hat w_{k}^t = w^t$) are received from client $k$ , the model update after local training $w_{k}^t$ can be recovered in the server as follows:
\begin{equation}
    w_{k}^t = r_{k}^{t} + \hat w_{k}^t = r_{k}^{t} + w^t
\end{equation}
where $\hat w_{k}^t$ is the global model in the last round. 
Similarly, if we predict the global model update in the client, through sharing predictors and uncompressed residuals, the aggregated model can also be recovered in the client.

\subsection{Model Trajectory Synchronization}
\label{sec:sync}

Since for the model residuals lossy compression is applied, i.e. $ \bar r \neq r$, the updated model of a sender $w$ can be recovered in receivers as $\hat w = \bar r + \tilde w $. Therefore, we say that the $w$ cannot be fully recovered in receivers, as $\hat w \neq w $. If we cache the original models $w$ in the sender and $\hat w$ in receivers, the trajectories in the sender and receiver are different, which leads to drift in results of shared predictors.

To avoid the drift effect, we synchronize the model trajectories in the sender by simulating the recovering process: The originally updated models are not cached in the trajectories; instead, we recover the model from compressed model residuals locally, in order to enforce the trajectories in the sender in the same way  as the trajectories in the receiver. The \myResFed pseudocode is given in Algorithm~\ref{alg:FedRes}.

\section{Experiment}
\label{sec:experiment}

\subsection{Experimental Settings}
Guided by the previous work by~\cite{caldas2018leaf}\cite{Li2020Fair}, we process and distribute the datasets MNIST~(\cite{mnist-2010}), Fashion-MNIST~(\cite{xiao2017/online}), SVHN~(\cite{netzer2011reading}), CIFAR-10~(\cite{Krizhevsky09learningmultiple}) and CIFAR-100~(\cite{netzer2011reading}) 
on a set of clients, and train LeNet-5, CNN\footnote{It consists of 5 convolutional and 3 fully connected layers.}, ResNet-18 on those federated dataset distributively, as shown in Tab.~\ref{table:exp_settings}. We provide details on the experimental settings in Sec.~\ref{app:dist_exp}.

\subsection{Residuals vs Gradients and Weights}
\label{sec:4.1}

\begin{figure*}[t]
    \includegraphics[trim=0 0 0 0,clip,width=1\linewidth]{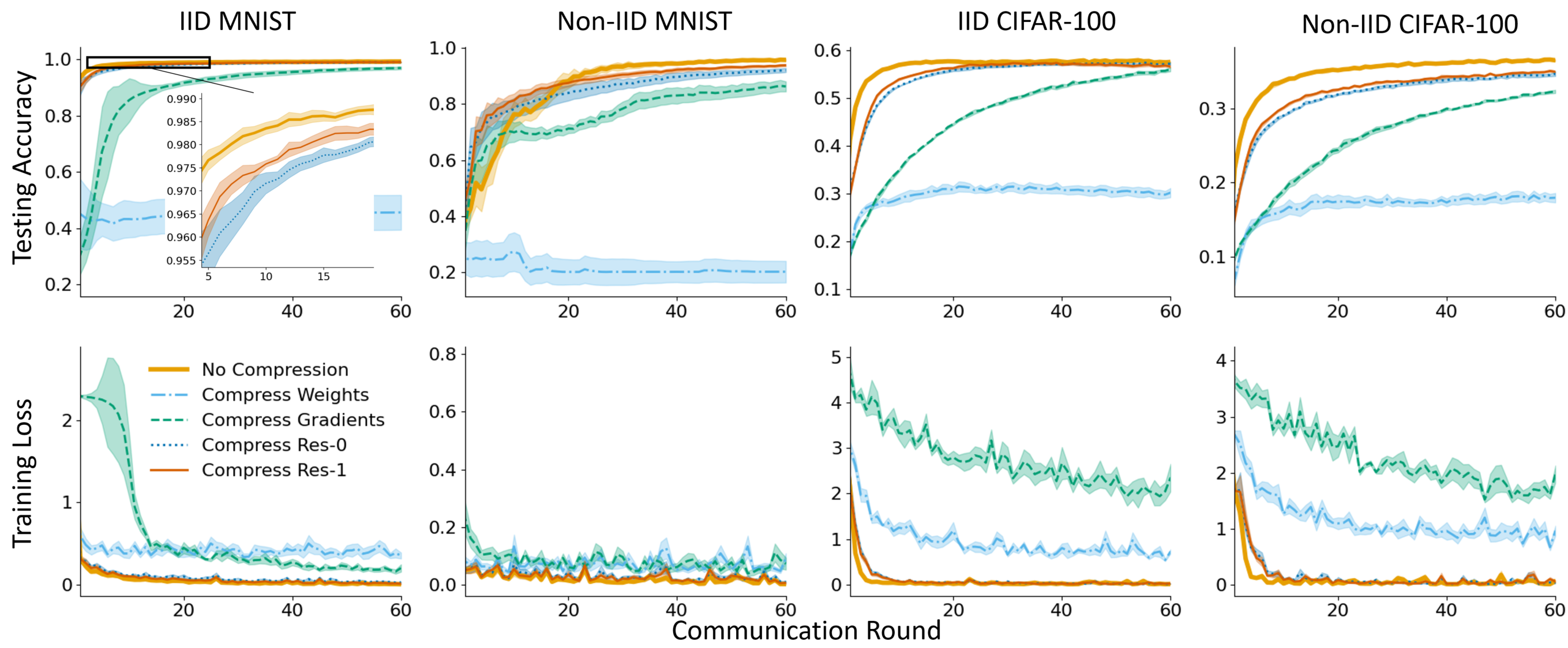}
    \caption{Comparison of compressing weights, gradients, residuals with stationary (Res-0) and linear (Res-1) predictors. The sparsities are 0.2 and 0.01 for training on MNIST and CIFAR-100 distributed in 10 clients, respectively. We quantize the non-zero parameters and use 1~bit to represent each of them. For Non-IID setting, each client owns only the data with half classes.}
    \vspace{0cm}
    \label{fig:compression}
\end{figure*}

On top of the basic evaluation in Fig.~\ref{fig:param_res}, we believe deep residual compression can save more communication volume in federated learning with minimum impact on the accuracy. 
Thus, we demonstrate the federated learning integrating compressing weights, gradients and two different residuals, i.e. stationary and linear residuals in Eq.~\ref{prediction}. 
As shown in Fig.~\ref{fig:compression}, the testing accuracy on both IID and Non-IID datasets from deep residual compression always outperforms weight and gradient compression. 
Also, the linear residuals can achieve a higher accuracy and faster convergence than stationary residuals. 
The results indicate communicating residuals in federated learning can enable larger compression ratio per communication round, compared to communicating other parameters.

\subsection{Communication Efficiency Improvement}
\label{sec:4.2}

\setlength{\tabcolsep}{3pt}
\begin{table*}[!t]
\centering
\fontsize{9}{12}\selectfont
\begin{threeparttable}
\caption{
The communication volume (CV) and the bitsaving rate (BR) to reach the target accuracy (ACC) for only using \myResFed in uploading (UL) and downloading (DL). We use FedAvg for both baseline (without any compression) and ResFed (Res-1).} Note that the compression ratio per communication round is set from $350\times$ to $375\times$. More details on experiment setup are shown in Sec.~\ref{app:dist_exp}.
\label{table:comm_eff}
\begin{tabular}{C{1.6cm}C{0.8cm}|C{1.4cm}|C{1.5cm}|C{1.5cm}|C{1.5cm}|C{1.5cm}|C{1.5cm}}
\toprule 
\multicolumn{2}{c|}{\multirow{2}{*}{Dataset}} & \multicolumn{2}{c|}{Fashion-MNIST} & \multicolumn{2}{c|}{CIFAR-10} & \multicolumn{2}{c}{SHVN}  \\
 \cmidrule(lr{0em}){3-8}
 & & IID & Non-IID & IID & Non-IID & IID & Non-IID \\
 \midrule 
\multicolumn{2}{c|}{Target ACC} & \multicolumn{2}{c|}{85\%}  & \multicolumn{2}{c|}{70\%} & \multicolumn{2}{c}{88\%} \\
\midrule 
\midrule 
\multirow{1}{*}{Baseline} & CV & 17.73 Mb & 29.55 Mb & 261.16 Mb & 587.61 Mb & 7.15 Gb & 10.73 Gb \\
\midrule 
\multirow{2}{*}{ResFed UL} & CV & 0.16 Mb & 0.28 Mb & 4.09 Mb & 6.79 Mb & 0.08 Gb & 0.17 Gb \\
& BR & \textbf{99.10\%} & \textbf{99.10\%} & \textbf{98.43\%} & \textbf{98.84\%} & \textbf{98.89\%} & \textbf{98.42\%} \\
\midrule 
\multirow{2}{*}{ResFed DL} & CV & 0.10 Mb & 0.21 Mb & 1.48 Mb & 4.61 Mb & 0.07 Gb & 0.11 Gb \\
& BR & \textbf{99.43\%} & \textbf{99.30\%} & \textbf{99.43\%} &  \textbf{99.22\%} & \textbf{99.02\%} & \textbf{98.97\%} \\
\bottomrule 
\end{tabular}
\end{threeparttable}
\end{table*}

Next, we evaluate the required communication volume for training three sizes of models on different datasets in both IID and Non-IID settings. Tab.~\ref{table:comm_eff} shows that to reach a promising target accuracy, \myResFed with lossy compression (compression ratio is set on $350\times-375\times$) can save on average around 99\% of the total communication volume for all clients in only up- or down-streaming. Furthermore, the bitsaving ratios of \myResFed on IID and Non-IID settings are similar, which indicates the compression performance of \myResFed is robust to data heterogeneity in federated learning. We show testing accuracy and training loss change with increasing required communication volume in \myResFed in Fig.~\ref{fig:tab1}. The results indicate communicating residuals in federated learning can remarkably save overall communication volume.

\subsection{Scalability for Resource-constrained Communication Environments}
\label{sec:4.3}

\begin{figure*}[t]
\centering
\begin{subfigure}[b]{0.31\textwidth}
   \includegraphics[trim=0 0 0 0,clip,width=1\linewidth]{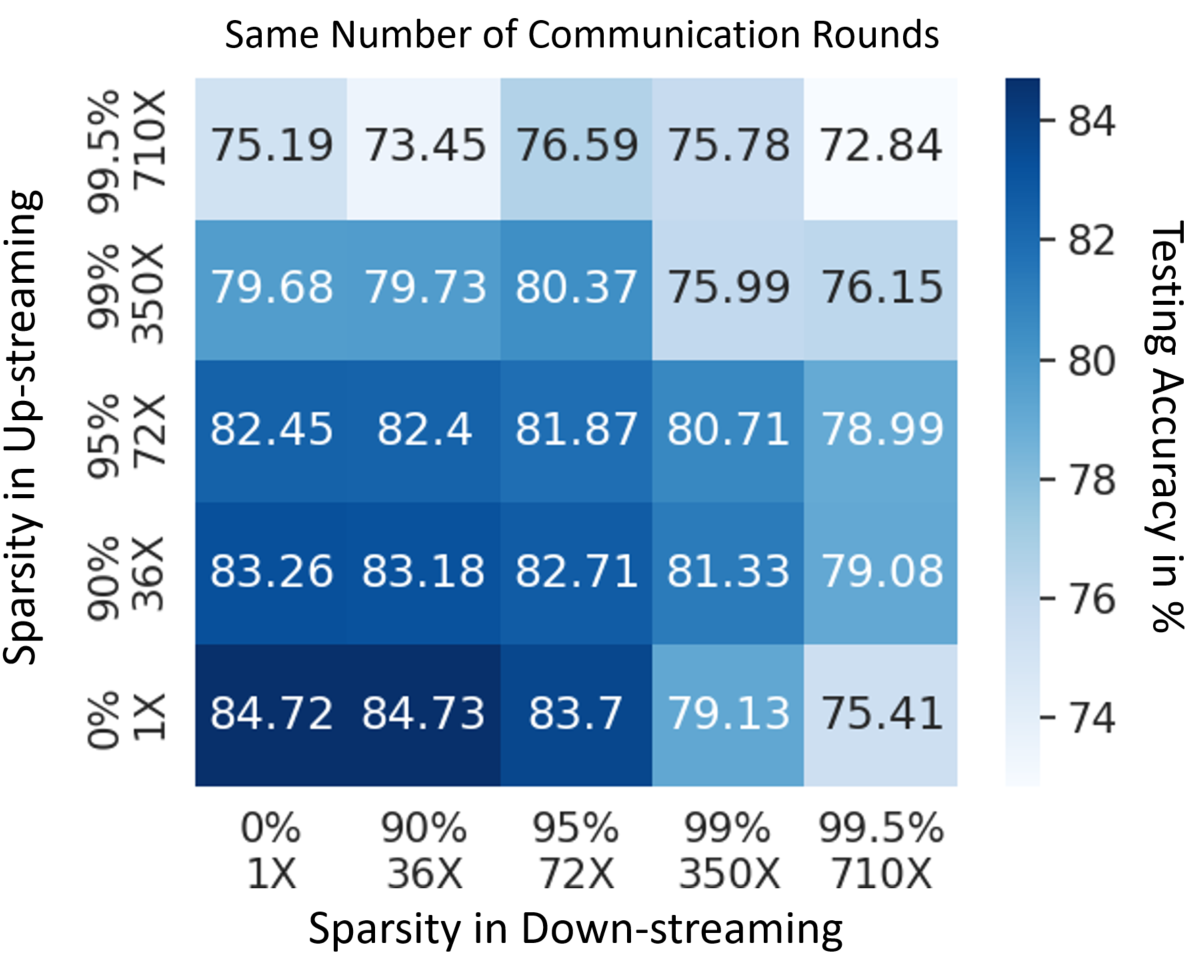}
   \caption{Number of communication rounds (300) is the same.}
   \label{fig:ResFed_sameround} 
\end{subfigure}
\hspace{0.2cm}
\begin{subfigure}[b]{0.62\textwidth}
  \includegraphics[trim=0 0 0 0,clip,width=1\linewidth]{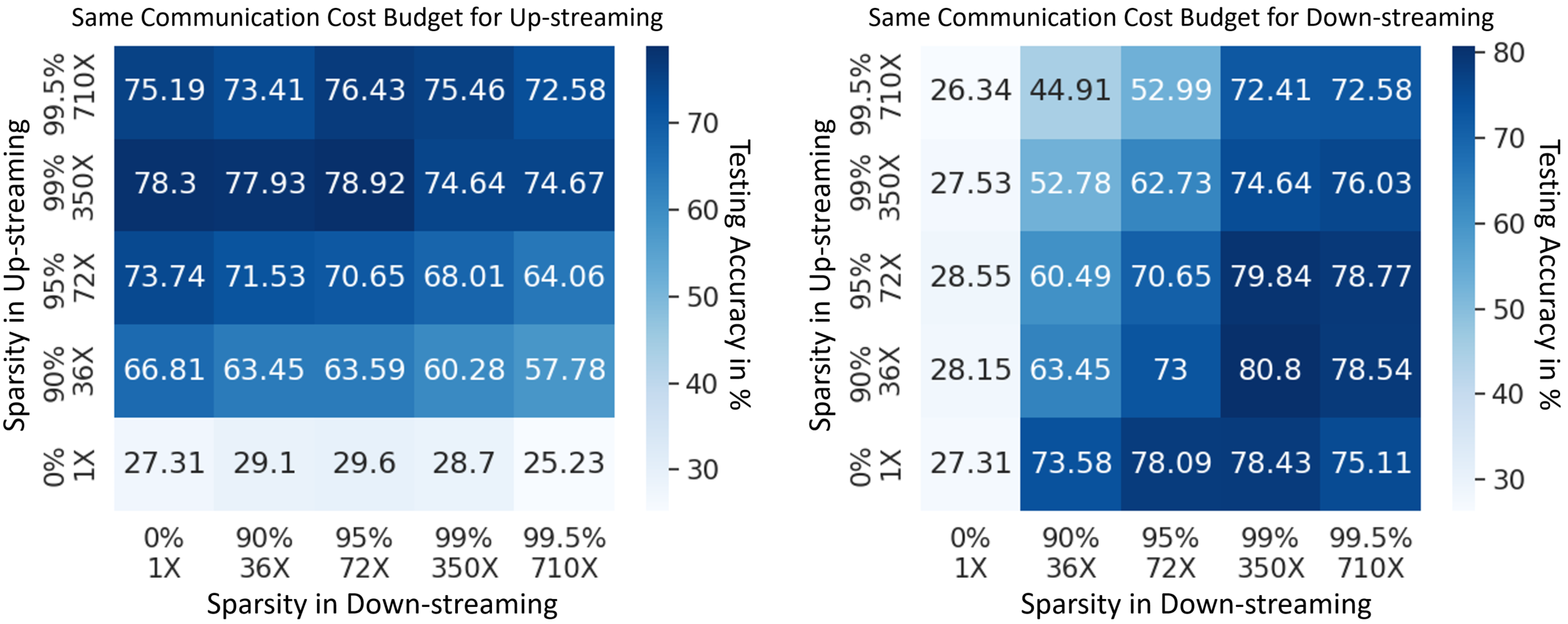}
   \caption{The communication cost budget (14 Mb) is the same in up-streaming (left) and down-streaming (right).}
   \label{fig:ResFed_samebudget}
\end{subfigure}
\caption{Testing accuracy on various values of sparsity and compression ratio per communication round in deep residual compression for both up- and down- streaming in \myResFed~(Res-1). We train a CNN model on a federated CIFAR-10 dataset with 10 clients. The testing accuracy decreases with a higher sparsity at the same number of communication rounds, while when the communication resource is constrained, the deep residual compression with higher sparsity can achieve more compromising testing accuracy.}
\label{fig:scalability}
\end{figure*}

Finally, we explore the scalability of \myResFed by tuning compression ratios for client-to-server and server-to-client, as in real application scenarios. The available network resources for up- and down-streaming can be heterogeneous. Fig~\ref{fig:scalability} shows the test accuracy effected for different values of sparsity, which leads to various compression ratios for each communication round for up- and down-streaming. From Fig.~\ref{fig:ResFed_sameround}, we can observe the testing accuracy reduces with higher compression ratio per communication round, when the number of communication rounds is always the same, i.e. set to 300. However, when we consider the dedicated budget for communication costs in up- or down-streaming, a large compression ratio in \myResFed can achieve better accuracy, as shown in Fig.~\ref{fig:ResFed_samebudget}. By adapting the compression ratio in resource-constrained communication environments, \myResFed can effectively enhance the federated learning using deep residual compression.

\section{Related Work}
\label{sec:related_work}

\mypara{Deep compression.}
Deep compression was originally proposed by~\cite{han2015deep_compression} and aims at compressing deep learning models by a pipeline including sparsification (pruning), quantization and encoding for more efficient deployment. 
Based on deep compression, \cite{lin2018deep} have proposed deep gradient compression to reduce the communication costs in distributed learning by compressing gradients rather than model weights, which can also be used in federated learning.

However, models are usually trained more than one epoch locally in federated learning~(\cite{pmlr-v54-mcmahan17a,li2020federated,wang2020tackling, karimireddy2020scaffold}), which results in gradient accumulation instead of gradients in other distributed learning scenarios~(\cite{sattler2019sparse}). Our conducted experiments also show compressing residuals can achieve better communication efficiency comparing to compress gradients due to the additional prediction step.
In \myResFed, we especially consider residuals, which eliminate the model similarity in a single inter-round of federated learning communication and achieve a better compression performance by leveraging the deep residual compression.

\mypara{Federated learning and communication efficiency.}
Communication efficiency is the key for deploying federated learning in real application scenarios, especially to train a large model. 
Previous research by~\cite{Yuan2020, karimireddy2020scaffold, hamer2020fedboost} attempted to reduce the number of needed communication rounds for a better communication efficiency.
Meanwhile, the proposed approaches by \cite{xu2020ternary, reisizadeh2020fedpaq, honig2022dadaquant} are built upon deep compression and focus on improving communication efficiency by decreasing the communication volume.
However, unlike compressing residuals in \myResFed, they compressed model weights without consideration of any potential redundancy in sequential updating of federated learning.
The recent work by~\cite{predcoding} has also mentioned the predictive model update in federated learning, which is concurrent to our work, but the information in the history of model updating is not considered for reducing the parameter redundancy there.

Additionally, all those algorithms above can only be used to improve the communication efficiency for up-streaming, while \myResFed can handle with up- and/or down-streaming for heterogeneous resource-constrained environments.
%


\mypara{Residuals in video encoding.}
The residuals have been widely and successfully utilized in video encoding since H.261~(\cite{girod1995comparison, li2021deep}).
By considering inter-frame correlations, the pixel values in the current frame are predicted from history frames and then only residuals, i.e. the deviations between predicted and the actual pixel values in the current frame, are encoded and streamed to the receivers. Inspired by the residuals in video encoding, we integrate the model residuals into federated learning in \myResFed, where the inter-round similarity of a model update is analogous to inter-frame correlation in video encoding.

\section{Conclusion}
\label{sec:conclusion}

In this work, we introduce a residual-based federated learning framework, which allows clients and the server to share residuals instead of weights or gradients. 
It achieves more efficient communication for both up- and down-streaming in federated learning by leveraging deep residual compression, and hence can be flexibly deployed in heterogeneous network environments.
Our conducted experimental evaluation shows that the framework remarkably reduces overall communication volume to reach the same prediction accuracy in standard federated learning. Compared to compressing model weights or gradients, \myResFed achieves higher accuracy and faster convergence speed.
%
%

\mypara{Limitations.}
We cache the recovered models as local and global trajectories for continual model prediction and residual computing in all clients and server. Assuming that we perform \myResFed with $N$ clients for training a model with $V$ 32-bit float parameters and we set the trajectory length on $T$, each client should use $2*32*V*T$ bits memory to cache the $2$ trajectories. Thus, the additional required memory size in each client is proportional to $T*V$. For the server, it needs $2*32*V*T*N$ bits memory to cache the local and global trajectories for all clients. In order to reduce the required memory, a potential solution is to cache the compressed models in the trajectories for both sender and receiver symmetrically, after model recovery. However, the accuracy of model prediction based on compressed trajectories is reduced, then the memory-accuracy trade-off needs to be investigated in future work.


\section{Reproducibility statement}

We provide the source-code for the implementation and evaluation of our proposed framework \myResFed in the code appendix. The user guidelines for installation and execution is given in the file \emph{README.md}. The details on our conducted experiments are provided in Sec.~\ref{app:dist_exp}. The source-code will be made publicly available after double-blind review.
\bibliography{iclr2023_conference}
\bibliographystyle{iclr2023_conference}
\newpage

\appendix
\section{Tables of Notations}
\label{subsec:A}

We provide an overview of the most relevant notations in Tab.~\ref{table:notation}.

\setcounter{table}{0}
\renewcommand{\thetable}{A\arabic{table}}

\begin{table*}[h]
\centering
\begin{threeparttable}
\caption{\centering Summary of mainly used notations in this paper.}
\label{table:notation}
\begin{tabular}{M{1.5cm}L{9cm}M{2cm}}
   \toprule 
    \centering\textbf{Notation} & \centering\textbf{Meaning}  & \textbf{Navigation}\\
    \midrule 

    $i$  & Index of clients & Sec.~\ref{sec:update} \\
    $k$  & Index of dedicated client & Sec.~\ref{sec:update}\\
    $t$  & Index of communication round & Sec.~\ref{sec:update}\\
    $\eta$  & Learning rate & Sec.~\ref{sec:sharing}\\
    $N$  & Number of connected clients & Sec.~\ref{sec:method}\\
    $T$  & Number of total communication rounds & Sec.~\ref{sec:trajectory}\\
    $V$  & Number of parameters in a machine learning model & Sec.~\ref{sec:conclusion}\\
    $w$  & Model weights & Sec.~\ref{sec:update}\\
    $w^t$  & Model weights in communication round $t$ & Sec.~\ref{sec:update}\\
    $w_k$  & Model weights in client $k$ & Sec.~\ref{sec:update}\\
    $\bar w$  & Compressed model weights & Sec.~\ref{sec:compression}\\
    $\tilde w$  & Predicted model weights & Sec.~\ref{sec:prediction}\\
    $\hat w$  & Received model weights for update & Sec.~\ref{sec:update}\\
    $g$  & Model gradients & Sec.~\ref{sec:sharing}\\
    $r$  & Model residuals & Sec.~\ref{sec:residual}\\
    $r^t$  & Model residuals in communication round $t$ & Sec.~\ref{sec:residual}\\
    $r_k$  & Model residuals in client $k$ & Sec.~\ref{sec:residual}\\
    $r_{i,dl}$  & Model residuals of client $i$ for uploading in client $i$ & Sec.~\ref{sec:residual}\\
    $r_{i,ul}$  & Model residuals of server for downloading in client $i$ & Sec.~\ref{sec:residual}\\
    $\bar r$  & Compressed model residuals & Sec.~\ref{sec:compression}\\
    $\hat r$  & Received model residuals  & Sec.~\ref{sec:update}\\
    
    $\mathcal{D}$  & Data set & Sec.~\ref{sec:update}\\
    $\mathcal{G}$  & Global trajectory queue & Sec.~\ref{sec:trajectory}\\
    $\mathcal{L}$  & Local trajectory queue & Sec.~\ref{sec:trajectory}\\
    \bottomrule 
\end{tabular}
\end{threeparttable}
\end{table*}

\newcommand{\vecline}{~~~\rule[.5ex]{2em}{0.4pt}~~~}

\section{Experimental Details and Further Results}
\label{app:dist_exp}

In this section, we provide the details on our conducted experiment in~\ref{sec:experiment}. We run on a computer cluster with 4$\times$~NVIDIA-A100-PCIE-40GB GPUs and 4$\times$~32-Core-AMD-EPYC-7513 CPUs. The environment is a Linux system with Pytorch 1.8.1 and Cuda 11.1.

We demonstrate the learning task on 5 different datasets:
\begin{itemize}
    \item MNIST~\cite{mnist-2010}: 60000 data points in the training set and 10000 data points in the test set. Each data point is a 28x28 gray-scale digit image, associated with a label from 10 classes.
    \item CIFAR-10~\cite{Krizhevsky09learningmultiple}: 50000 data points in the training set and 10000 data points in the test set. Each data point is a 32x32 RGB image, associated with a label from 10 classes.
    \item Fashion-MNIST~\cite{xiao2017/online}: 60000 data points in the training set and 10000 data points in the test set. Each data point is a 28x28 gray-scale image, associated with a label from 10 classes.
    \item SVHN~\cite{netzer2011reading}: 73257 data points in the training set and 26032 data points in the test set. Each data point is a 32x32 RGB digit image, associated with a label from 10 classes.
    \item CIFAR-100~\cite{netzer2011reading}: 50000 data points in the training set and 10000 data points in the test set. Each data point is a 32x32 RGB image, associated with a label from 100 classes.
\end{itemize}
The models trained on those dataset are shown in Tab.~\ref{table:exp_settings}.

\setcounter{table}{0}
\renewcommand{\thetable}{B\arabic{table}}
\setlength{\tabcolsep}{3pt}
\begin{table*}[h]
\centering
\fontsize{9}{12}\selectfont
\begin{threeparttable}
\caption{Dataset and models in experiments}
\label{table:exp_settings}
\begin{tabular}{C{1.5cm}|C{2.1cm}C{2.1cm}C{2.1cm}C{2.1cm}C{2.1cm}}
\toprule 
Dataset & MNIST & CIFAR-10 & Fashion-MNIST & SVHN & CIFAR-100 \\
\midrule 
Model & LeNet-5 & CNN & LeNet-5 & ResNet-18 & CNN\\
\midrule 
\# of Param & 61706 & 1020160 & 61706 & 11175370 & 1020160\\
\midrule 
Size in MB & 0.25 & 4.08 & 0.25 & 44.70 & 4.08\\
\bottomrule 
\end{tabular}
\end{threeparttable}
\end{table*}

\subsection{Experiments for Sec.~\ref{sec:4.1}}

We divide the dataset, i.e. MNIST and CIFAR-100, into 10 clients and run the FedAvg with local optimizer of stochastic gradient descent (SGD) (momentum is 0.9) to train LeNet-5 and CNN (with 5 convolutional and 3 fully connected layers), respectively.  The learning rate is fixed on 0.01 and the batch size is 64 for all tests. In Non-IID data setting, each client owns only 2 out of 10 classes in MNIST and 50 out of 100 classes in CIFAR-100. 

In clients, we consider 5 different approaches as follows:
\begin{itemize}
    \item No Compression: The standard federated learning without any compression methods is used as the baseline, which provides the best results among all of the approaches, when number of communication rounds is the same.
    \item Compress Weights: Before communication in standard federated learning, the model weights are first compressed.
    \item Compress Gradients: The gradients in each epoch are compressed and communicated to the server.
    \item Compress Res-0: The residuals are computed by stationary predictor, i.e. Eq.~\ref{prediction} with $T=0$. 
    \item Compress Res-1: The residuals are computed by linear predictor, i.e. Eq.~\ref{prediction} with $T=1$.
\end{itemize}
For each approach on each dataset, we run 10 tests with different seeds and show the mean value and standard variance in Fig.~\ref{fig:compression}.

We compress the those model parameters using deep compression pipeline~(\cite{han2015deep_compression}) only for client-to-server. In particular, we set sparsity on 80\% and 99\% for residuals in LetNet and CNN, respectively. We use SGD optimizer momentum of 0.9. The number of local epoch is 1. Those sparsified parameters are zero-parameters and the number of the continually appearing zero-parameters are encoded as a 16-bit float parameters~(\cite{lin2018deep}). After that, we quantize the non-zero parameters in 1 bit with median value of positive and negative parameters~(\cite{xu2020ternary}). Finally, Huffman encoding~(\cite{van1976construction}) is used.

\subsection{Experiments for Sec.~\ref{sec:4.2}}

We train LeNet-5, CNN and ResNet18 (size from small to large) on Fashion-MNIST, CIFAR-10 and SVHN with 10 clients in both IID and Non-IID settings. We demonstrate the ResFed and lossy compress the residuals either only for uploading (UL) or downloading (DL) to study the effects on each direction separately. The learning rate is fixed on 0.01 and the batch size is 64 for all tests. In Non-IID data setting, each client owns 50\% classes (5 out of 10). We use mean values from 5 tests in each experiment for the evaluation in Tab.~\ref{table:comm_eff} and show the results with standard variance in Fig.~\ref{fig:tab1}, which indicate the overall communication volume can be remarkably reduced in \myResFed.

\setcounter{figure}{0}                       
\renewcommand\thefigure{B.\arabic{figure}}

\begin{figure*}[t]
\centering
\begin{subfigure}[b]{1\textwidth}
   \includegraphics[trim=0 0 0 0,clip,width=1\linewidth]{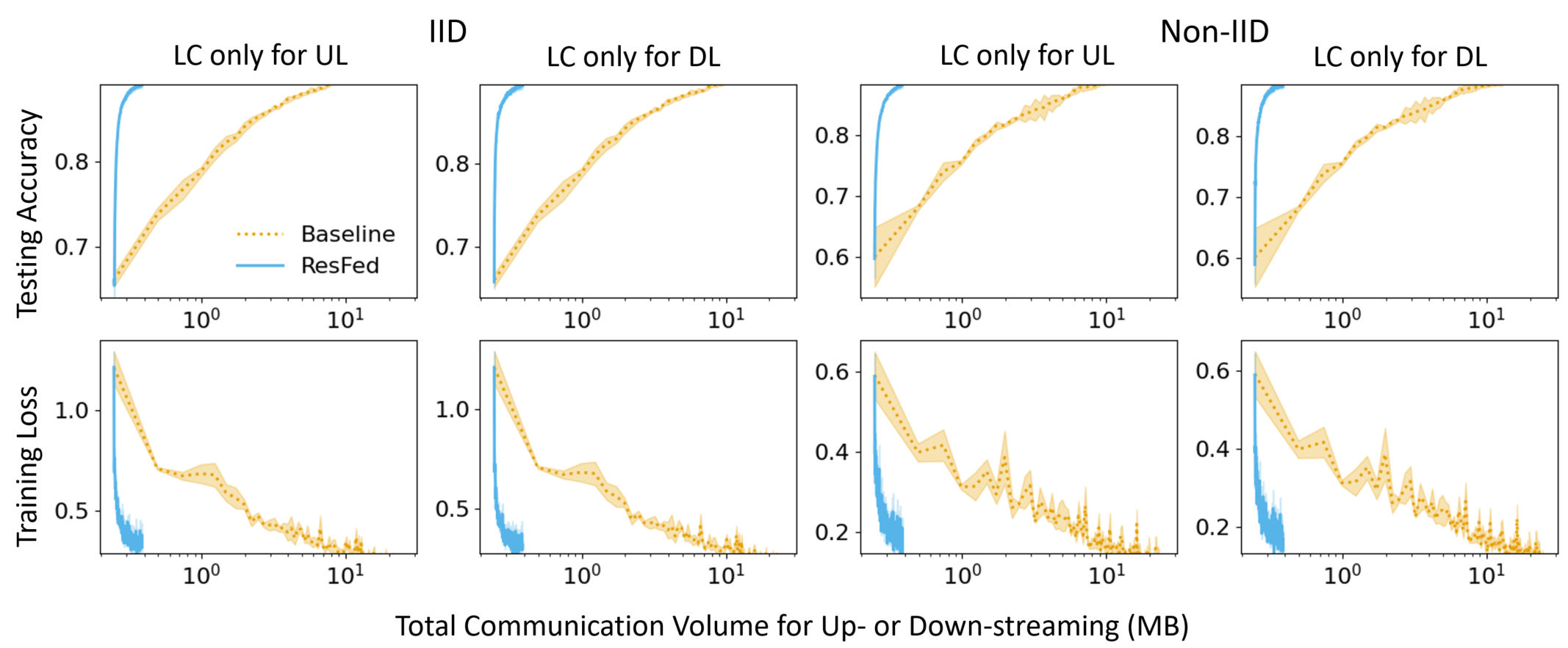}
   \caption{Training LeNet-5 on Fashion-MNIST.}
   \label{fig:tab1_fashionmnist} 
\end{subfigure}
\hspace{0.1cm}
\begin{subfigure}[b]{1\textwidth}
  \includegraphics[trim=0 0 0 0,clip,width=1\linewidth]{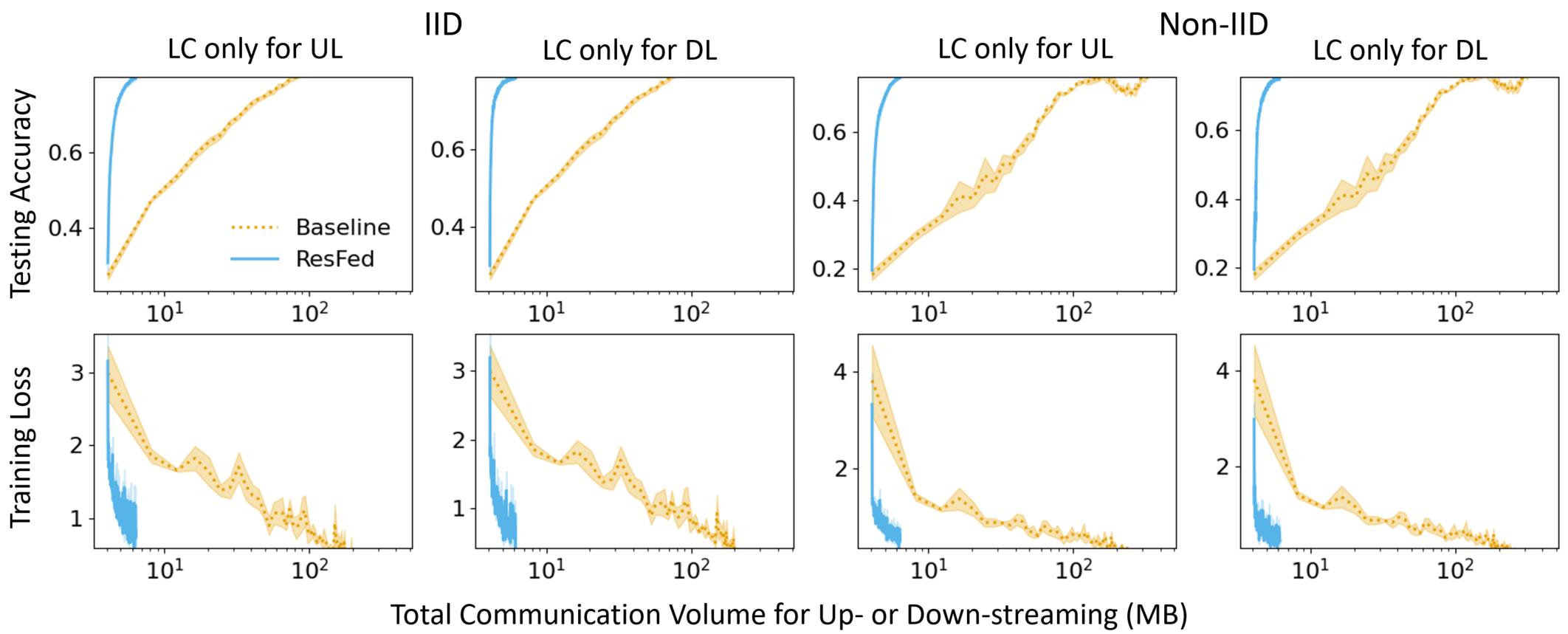}
   \caption{Training CNN on CIFAR-10.}
   \label{fig:tab1_cifar10}
\end{subfigure}
\hspace{0.1cm}
\begin{subfigure}[b]{1\textwidth}
  \includegraphics[trim=0 0 0 0,clip,width=1\linewidth]{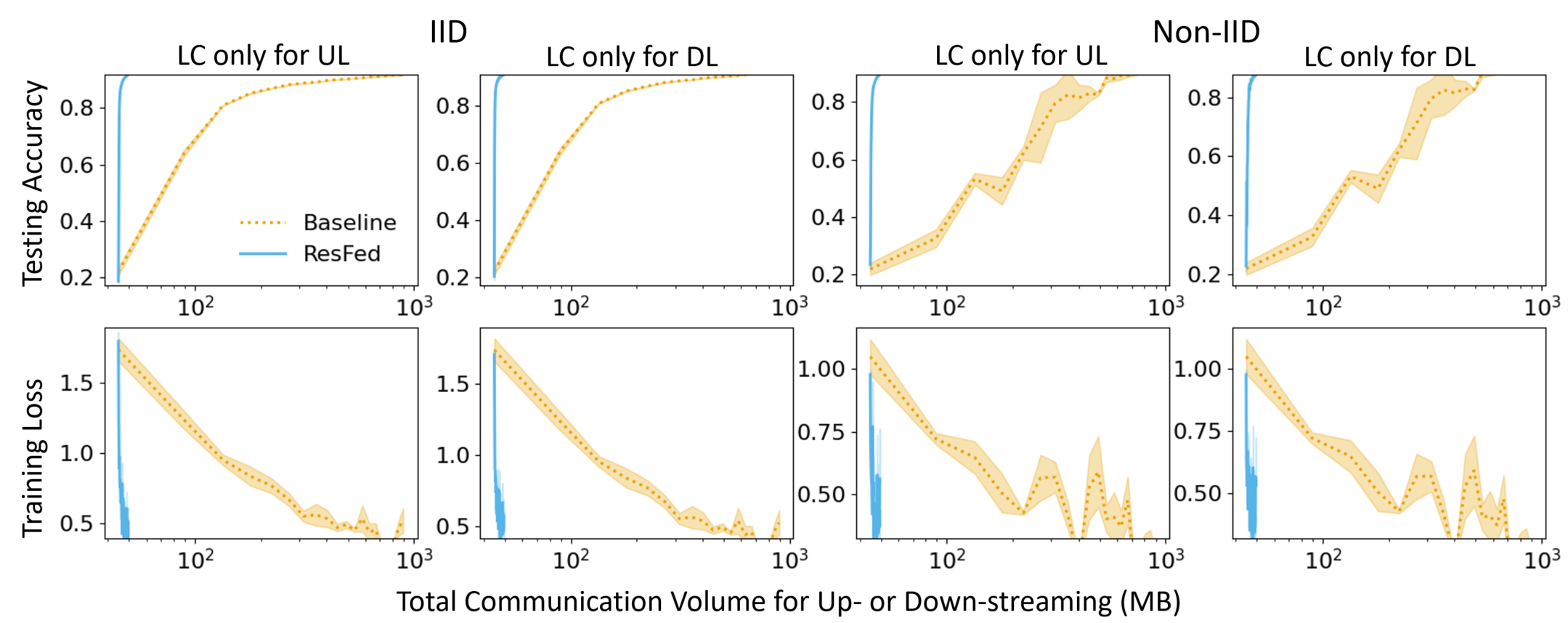}
   \caption{Training ResNet-18 on SVHN.}
   \label{fig:tab1_svhn}
\end{subfigure}
\caption{Overall communication efficiency enhanced by \myResFed with Lossy Compression (LC) for only Uploading (UL) or Downloading (DL)}
\label{fig:tab1}
\end{figure*}

For all experiments, we set the sparsity on 99\% and use 1 bit for each non-zero parameters. Consequently, the compression ratio per communication round for LetNet-5 is about $350\times$, CNN is about $375\times$ and ResNet-18 is about $356\times$.

\subsection{Experiments for Sec.~\ref{sec:4.3}}

To show the correlation between deep residual compression in up- und down-streaming, we train the CNN model on IID CIFAR-10 with 10 clients and tune the sparsity for realizing different compression ratios per communication round in \myResFed. The learning rate is fixed on 0.01 and the batch size is 64. We use SGD optimizer momentum of 0.9. The number of local epoch is 1. Specifically, the value of sparsity is $\{0\%, 90\%, 95\%, 99\%, 99.5\%\}$ for both up- and down-streaming and then set 1 bit for all non-zero parameters in quantization.

\section{Understanding Residuals}

We provide an illustration of model transitions during federated learning in Fig.~\ref{fig:understanding}. 
Given a sender and a receiver (both can be a client or a server), the communication and operation result in model transition. 
Note that for a client, the operation is local training; for a server, the operation is aggregation.
We consider the model transition caused by an operation as a internal model transition, and by communication as a external model transition. Then, as explained in Sec.~\ref{sec:update}, the model is updated twice between two communication rounds, which is shown in Fig.~\ref{fig:understanding} as dual model transition. Consequently, we can have an internal and an external model transition trajectory in both sender and receiver. 
Note that for a client, the internal model transition trajectory is a local model trajectory; for a server, the internal model transition trajectory is a global model trajectory, as described in Sec.~\ref{sec:trajectory}.

\label{sec:understanding}

\setcounter{figure}{0}                       
\renewcommand\thefigure{C.\arabic{figure}}

\begin{figure*}[ht]
\includegraphics[trim=0 0 0 0,clip,width=1\linewidth]{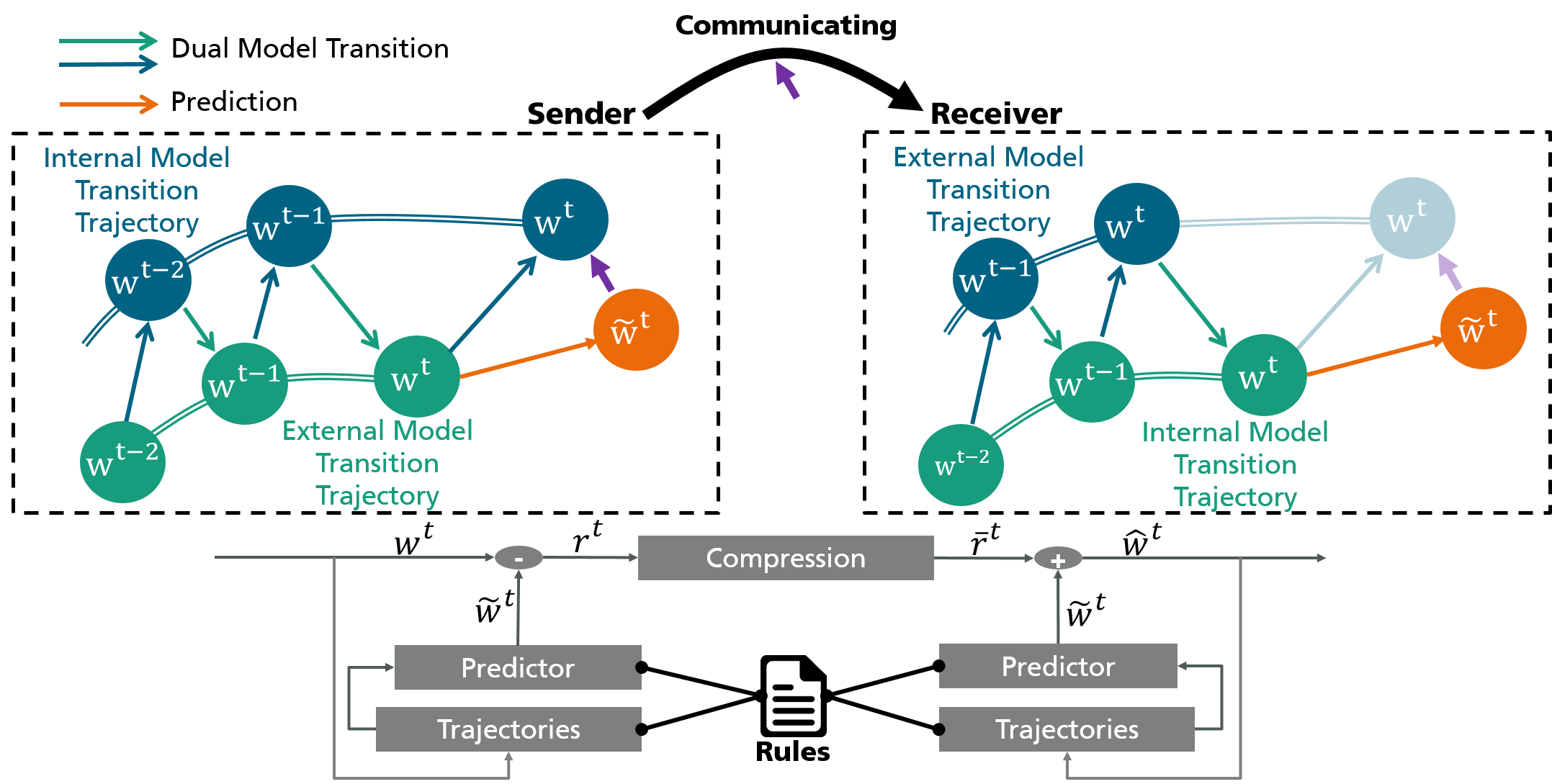}
\caption{Residual generation in \myResFed during model transitions. For a better overview, we simplify the system by disregarding the trajectory synchronization step in Sec.~\ref{sec:sync}.}
\vspace{0cm}
\label{fig:understanding}
\end{figure*}

\begin{figure*}[ht]
\includegraphics[trim=0 0 0 0,clip,width=1\linewidth]{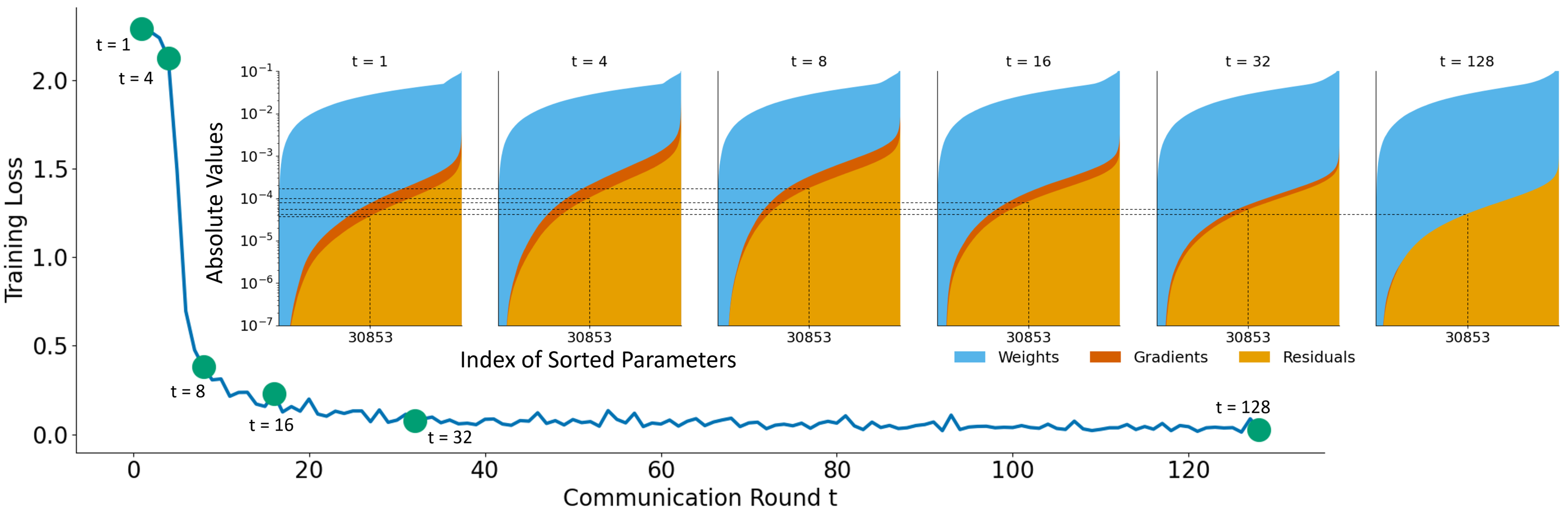}
\caption{Value comparison of model parameters, gradients and residuals in federated learning. We train a LeNet with $61706$ weights of 32-bits float on MNIST distributed among $10$~clients, with fixed learning rate $0.001$ and batch size $64$. For fairly comparing gradients and residuals, the number of local epoch in each client is set as $1$. We set 6 checkpoints when the number of communication rounds is $\{1, 4, 8, 16, 32, 128\}$. The results show that most values of residuals are smaller than weights and gradients during the training. It indicates that lossly compressing residuals naturally lose less information and have a smaller affect on the accuracy.} 
\vspace{0cm}
\label{fig:param_res}
\end{figure*}

Finally, \myResFed allows the sender to predict the model for the next internal model transition, which is shown in orange. Meanwhile the sender does the operation to execute the internal model transition and residuals (in purple) are deduced from the difference of both model transition results.
We believe the predicted model can be closer to the updated model than the previous model, which leads to smaller values of residuals. To evaluate it, we set 6 checkpoints when the number of communication rounds is $\{1, 4, 8, 16, 32, 128\}$, and show the values of model weights, gradients and residuals in Fig.~\ref{fig:param_res}. 
Based on this, the residuals can be compressed smaller than weights and gradients.

\end{document}